\documentclass[letterpaper, 10 pt, conference]{ieeeconf}
\usepackage{graphicx}
\IEEEoverridecommandlockouts
\usepackage{xcolor}
\usepackage{subcaption}
\usepackage{tabularray}
\usepackage{array}  
\usepackage{csquotes}
\usepackage{hyperref}
\usepackage{cite}  
\usepackage[labelsep=period]{caption}

\title{\LARGE \bf
Hierarchical Question-Answering for Driving Scene Understanding Using Vision-Language Models
}

\author{Safaa Abdullahi Moallim Mohamud\textsuperscript{1}, Minjin Baek\textsuperscript{1}, and Dong Seog Han\textsuperscript{2,*}%
\thanks{$^{1}$Center for ICT and Automotive Convergence, Kyungpook National University, Daegu, South Korea
        {\tt\small \{safaa, mbaek\}@knu.ac.kr}}
\thanks{\textsuperscript{2}School of Electronic and Electrical Engineering, Kyungpook National University, Daegu, South Korea
        {\tt\small dshan@knu.ac.kr}}%
\thanks{\textsuperscript{*}Corresponding author: Dong Seog Han.}
\thanks{This research was supported by Basic Science Research Program through the National Research Foundation of Korea (NRF) funded by the Ministry of Education (2021R1A6A1A03043144).}%
}

\usepackage{eso-pic}

\IEEEoverridecommandlockouts

\begin{document}

\AddToShipoutPicture*{%
  \AtTextUpperLeft{%
    \parbox{\textwidth}{%
      \footnotesize\centering
      \textbf{This work has been submitted to the IEEE for possible publication. 
      Copyright may be transferred without notice, after which this version 
      may no longer be accessible.}\vspace{1em}
    }
  }
}



\maketitle

\thispagestyle{empty}
\pagestyle{empty}

\begin{abstract} In this paper, we present a hierarchical question-answering (QA) approach for scene understanding in autonomous vehicles, balancing cost-efficiency with detailed visual interpretation. The method fine-tunes a compact vision-language model (VLM) on a custom dataset specific to the geographical area in which the vehicle operates to capture key driving-related visual elements. At the inference stage, the hierarchical QA strategy decomposes the scene understanding task into high-level and detailed sub-questions. Instead of generating lengthy descriptions, the VLM navigates a structured question tree, where answering high-level questions (e.g., \enquote{Is it possible for the ego vehicle to turn left at the intersection?}) triggers more detailed sub-questions (e.g., \enquote{Is there a vehicle approaching the intersection from the opposite direction?}). To optimize inference time, questions are dynamically skipped based on previous answers, minimizing computational overhead. The extracted answers are then synthesized using handcrafted templates to ensure coherent, contextually accurate scene descriptions. We evaluate the proposed approach on the custom dataset using GPT reference-free scoring, \textcolor{black}{demonstrating its competitiveness with state-of-the-art methods like GPT-4o in capturing key scene details while achieving significantly lower inference time.} Moreover, qualitative results from real-time deployment highlight the proposed approach's capacity to capture key driving elements with minimal latency. 
\end{abstract}

\section{Introduction}

Autonomous driving has gained significant attention in recent years, with advancements in artificial intelligence enabling vehicles to perceive and understand their surroundings \cite{lai2023sphericaltransformerlidarbased3d}. Scene understanding plays a crucial role in ensuring safe and efficient navigation, as it allows autonomous vehicles to analyze their environment and make informed decisions \cite{nie2024reason2driveinterpretablechainbasedreasoning}. Recent methods leverage large vision-language models (LVLMs) to enhance interpretability by generating natural language scene descriptions \cite{2023GPT4VisionSC}. 
However, deploying such models on real-world autonomous systems comes with major challenges.

One of the primary obstacles in leveraging large language models (LLMs) and LVLMs for scene understanding is their computational cost \cite{tian2024drivevlm}. These models require substantial hardware resources, making offline deployment on autonomous vehicles challenging. While cloud-based solutions offer an alternative, \textcolor{black}{their inherent network latency and reliability constraints pose challenges for time-sensitive decision-making and real-time processing in autonomous driving}. A potential solution to this problem is the use of compact vision-language models (VLMs), which offer a significant reduction in computational cost and inference time. \textcolor{black}{These models are effective} in generating short captions that briefly describe an image \cite{li2022blipbootstrappinglanguageimagepretraining}. However, they \textcolor{black}{often} struggle when tasked with producing longer, more detailed descriptions. 
This limitation makes them less effective in providing comprehensive scene understanding, as they fail to capture the full context of a driving scenario in a structured and interpretable manner.

\begin{figure*}[thpb]
  \includegraphics[width=\textwidth,height=10cm]{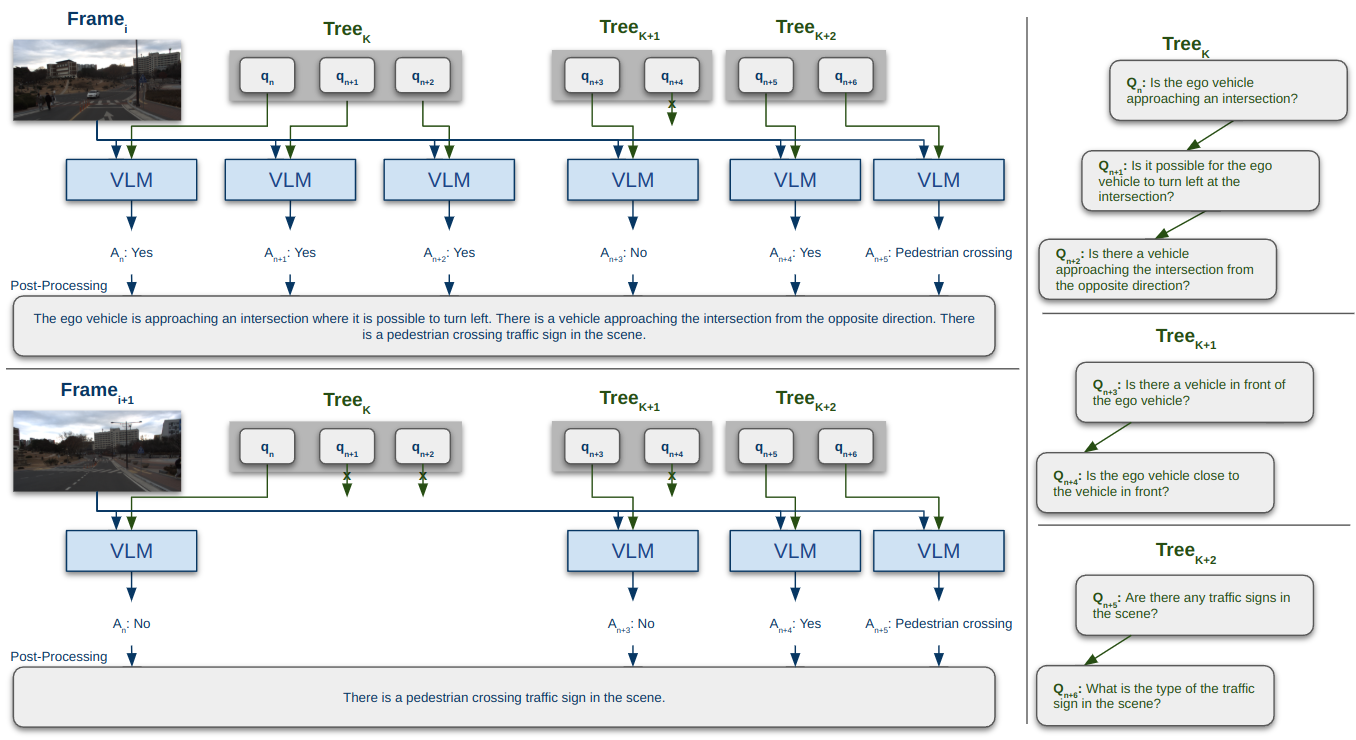}
  \caption{Overview of the hierarchical QA approach for scene understanding in autonomous vehicles. }
  \label{framework}
\end{figure*}

To address these challenges, we propose a lightweight hierarchical question-answering (QA) approach for real-time scene understanding in autonomous vehicles, designed to balance cost-efficiency with detailed visual interpretation. Our method fine-tunes a compact VLM on a custom dataset tailored to the specific geographical area where the autonomous vehicle operates. \textcolor{black}{The hierarchical QA framework enables multi-level reasoning, allowing the model to refine its understanding from broad scene descriptions to specific driving-relevant details}. By leveraging prior knowledge of the vehicle's \textcolor{black}{operating environment}, the custom dataset captures key driving-related visual elements, including \textcolor{black}{road geometry, intersections, vehicles, pedestrians, traffic signs, speed bumps, and crosswalks}. Furthermore, this approach is inherently scalable: by expanding the dataset to \textcolor{black}{cover diverse driving environments, such as dense urban centers and suburban roads, this methodology scales to larger regions} while maintaining efficiency.

To achieve both efficiency and accuracy in \textcolor{black}{driving} scene understanding, our approach utilizes a hierarchical QA strategy that decomposes the task into high-level and detailed sub-questions. Instead of relying on the VLM to generate lengthy descriptions, the approach navigates a structured question tree, where answering a high-level question (e.g., \enquote{Is it possible for the ego vehicle to turn left at the intersection?}) triggers more detailed sub-questions (e.g., \enquote{Is there a vehicle approaching the intersection from the opposite direction?}), as illustrated in Fig. \ref{framework}. \textcolor{black}{It is important to note that the figure serves as an example; the actual proposed tree contains more subtrees and generates longer, more detailed descriptions.} To minimize inference time, questions are skipped if their parent questions have already been answered with \enquote{no,} dynamically adjusting the number of questions for each frame based on the visual elements present in the image. Subsequently, the extracted answers are then synthesized using handcrafted templates, ensuring that scene descriptions remain coherent, contextually accurate, and linguistically structured. The primary objective is to provide \textcolor{black}{detailed and accurate scene descriptions of driving environments} while minimizing computational overhead and inference time. This balance of efficiency and quality makes our approach suitable for real-time autonomous driving applications.


The key contributions of our research are as follows:

\begin{itemize}

\item The proposed hierarchical QA approach leverages a compact VLM model fine-tuned on a custom dataset specific to the operating environment, capturing key driving-related visual elements.

\item 

Real-time scene understanding is achieved through a structured question tree that breaks down high-level questions into detailed sub-questions, \textcolor{black}{optimizing inference time and computational efficiency. With an average inference time of 423 ms, this approach supports real-time autonomous driving applications that involve high-level scene interpretation and decision-making.}


\item \textcolor{black}{The proposed approach generates driving scene descriptions evaluated using a reference-free GPT score, achieving a score of 65 out of 100, demonstrating its effectiveness in capturing key scene details compared to GPT-4o’s score of 77.} 


\end{itemize}

\section{Related Work}

\subsection{Scene Understanding for Autonomous Driving}

Scene understanding is a crucial task for autonomous vehicles to ensure safe and efficient operations. \textcolor{black}{Previously proposed} supervised 3D scene methods, such as Clip2scene \cite{chen2023clip2scenelabelefficient3dscene} and SphereFormer \cite{lai2023sphericaltransformerlidarbased3d}, have made notable progress in this area. 
Moreover, recent advancements have explored a range of methodologies, including the integration of VLMs, LVLMs and 3D scene recognition techniques. Notably, approaches such as OpenScene \cite{peng2023openscene3dsceneunderstanding} and PLA \cite{ding2023plalanguagedrivenopenvocabulary3d} combine 3D scene features with VLMs such as CLIP \cite{radford2021learningtransferablevisualmodels} to enhance model generalization. GPT-4V \cite{2023GPT4VisionSC} analyzes traffic accident scenarios and urban environments with potential risks. Similarly, other approaches \cite{9306804} use image captioning to describe traffic scenes, while AnomalyCLIP \cite{zanella2023delvingcliplatentspace} leverages CLIP for video anomaly detection, utilizing context optimization \cite{Zhou_2022} to identify abnormal events. Furthermore, models like LiDAR-LLM \cite{yang2023lidarllmexploringpotentiallarge} and Reason2Drive \cite{nie2024reason2driveinterpretablechainbasedreasoning} focus on complex decision-making frameworks, integrating LiDAR data and structured reasoning steps.  

While previous work has shown promising results in traffic scene understanding, these methods often involve complex, resource-intensive architectures that may not be scalable for real-time applications in autonomous driving. By contrast, our hierarchical QA approach, \textcolor{black}{combined with efficient post-processing, ensures scalability and resource efficiency, enabling low-latency inference and enhancing its suitability for deployment in autonomous driving systems.}

\subsection{LLMs in Autonomous Driving}

Recent advancements in LLMs and LVLMs have facilitated their integration into driving decision-making frameworks \cite{gao2020vectornetencodinghdmaps, hawke2021reimaginingautonomousvehicle}. However, despite these strides, a critical challenge remains in modern autonomous driving systems: the lack of interpretability in decision-making processes \cite{arrieta2019explainableartificialintelligencexai}. Understanding the rationale behind driving decisions is essential for identifying areas of uncertainty, building trust between humans and AI, and ensuring safe and effective human-AI collaboration \cite{article}. 

Furthermore, several studies have explored the use of LLMs in autonomous driving, either through prompt engineering to derive driving decisions from textual descriptions of the vehicle’s environment \cite{sha2023languagempclargelanguagemodels, shah2022lmnavroboticnavigationlarge, wen2024diluknowledgedrivenapproachautonomous} or by fine-tuning LLMs to predict future actions or plan vehicle trajectories \cite{chen2023drivingllmsfusingobjectlevel, mao2023gptdriverlearningdrivegpt}. DriveGPT4 \cite{xu2024drivegpt4interpretableendtoendautonomous} and RAG-Driver \cite{yuan2024ragdrivergeneralisabledrivingexplanations} fine-tuned multimodal LLMs on real-world driving videos to predict future throttle and steering angles. Similarly, DriveMLM \cite{wang2023drivemlmaligningmultimodallarge} and LMDrive \cite{shao2023lmdriveclosedloopendtoenddriving} incorporated camera data and ego-vehicle states from the CARLA simulator to generate driving instructions. These approaches demonstrate the potential of LLMs to improve decision-making by processing and interpreting both textual and visual information. \textcolor{black}{However, the inference delay of these methods poses challenges for their deployment in real-time autonomous driving systems.}

\section{Proposed Methodology}

The proposed methodology introduces a compact scene understanding system that leverages a hierarchical QA approach, with \textcolor{black}{Fig. \ref{framework} serving as an illustrative example.} A visual question answering (VQA) dataset is curated for the specific geographical region where the autonomous vehicle operates. Subsequently, a compact VLM is fine-tuned on this dataset to enhance its domain-specific understanding. At the inference stage, instead of generating free-form text, handcrafted templates are used to ensure coherence and contextual accuracy. The post-processing step converts positive answers into structured descriptions, which are sequentially compiled into a final scene interpretation. To optimize latency, the questions are organized into multiple decision trees, enabling the model to process only the relevant questions, rather than evaluating all possible queries. \textcolor{black}{This dynamic structure reduces inference time to a level suitable for deployment in driving scene understanding for autonomous systems, minimizing computational overhead while maintaining high accuracy.}

\subsection{Dataset Generation}

To create a dataset tailored to the autonomous vehicle’s operating environment, \textcolor{black}{camera images were collected using a front-facing camera (1920 × 750 resolution) mounted on the test vehicle, ensuring a clear view of the road ahead.} Frames were extracted at one frame per second, providing sufficient temporal coverage of the driving environment. Since our test vehicle operated exclusively within the university campus environment, the dataset was designed accordingly\textcolor{black}{---for example, omitting questions about traffic lights, which were absent in the vehicle’s region of operation}. \textcolor{black}{Following the extraction process,} a total of 465 frames were collected, each annotated with answers to 41 predefined questions, covering key driving elements such as road geometry, intersections, \textcolor{black}{vehicles, pedestrians}, traffic signs, speed bumps, and crosswalks. The annotation process was conducted manually, with a constrained set of possible responses, \textcolor{black}{such as yes, no, left, right, bicycle crossing, pedestrian crossing, and none.} The \enquote{none} label was assigned when a question was inapplicable due to prior responses---e.g., if no pedestrian was detected, subsequent questions \textcolor{black}{related to pedestrians} were marked as \enquote{none.}

The dataset encompasses a diverse range of driving scenarios, including intersections, pedestrian encounters, and road obstructions. Specifically, challenging situations such as unprotected left turns, pedestrians on the road, vehicle blockages, and vehicle merging are incorporated to ensure comprehensive model evaluation, as shown in Table \ref{Dataset_Summary}. \enquote{Comprehensive Area Exploration} refers to driving across \textcolor{black}{the vehicle’s region of operation} using multiple routes to capture diverse driving scenarios. The dataset is split into 80\% for training (15,006 question-answer pairs) and 20\% for validation (4,059 question-answer pairs), providing a structured foundation for fine-tuning the vision-language model.


\begin{table}[h]
\caption{Summary of the Proposed Dataset}
\label{Dataset_Summary}
\begin{center}

\begin{tabular}{c|c|c|c}
Scenario & Frames  & Train &Val\\
\hline
Comprehensive Area Exploration	&	250	&75\% & 25\%  \\
Unprotected Left Turn	&117		&80\% & 20\% \\
Pedestrian on the Road &	60		&80\% & 20\% \\
Vehicle Blocking the Road	&	20		&100\% &0\%\\
Merging Traffic	&18		&100\% & 0\%\\
\end{tabular}
\end{center}
\end{table}

\subsection{VLM Fine-Tuning and Hierarchical QA Inference}

For the fine-tuning and inference stages, a compact VLM is employed. All trainable parameters are fine-tuned on the collected dataset to ensure efficient adaptation. The fine-tuned weights are chosen based on the highest VQA accuracy achieved on the validation set. A hierarchical questioning technique is used to mitigate the limitations of smaller VLMs in generating long paragraphs. Instead of generating a single long paragraph, the system asks structured questions in a hierarchical manner, enabling efficient processing and resource minimization. This allows the VLM to generate meaningful scene descriptions without compromising speed or scalability. The hierarchical QA strategy optimizes inference time by dynamically selecting relevant questions based on the visual elements in the scene. For example, if the question \enquote{Is the ego vehicle moving on a straight road?} is answered affirmatively, subsequent questions such as \enquote{In which direction does the road curve?} are automatically answered as \enquote{none,} reducing unnecessary computations. Similarly, if the answer to \enquote{Is it possible for the ego vehicle to turn right at the intersection?} is \enquote{no,} other related questions, such as \enquote{Is there a vehicle approaching the intersection from the opposite direction?} are skipped.



\subsection{Structured Post-Processing}

During the inference post-processing phase, handcrafted templates are used to transform the model’s answers into structured scene descriptions. When a question is answered affirmatively, its corresponding template-generated sentence is appended to the final scene description. For example, if the hierarchical QA mechanism determines that the ego vehicle is moving on a curved road, the post-processing step generates the sentence: \enquote{The ego vehicle is moving on a curved road.} If a subsequent question confirms that the ego vehicle is approaching an intersection, the description is extended to: \enquote{The ego vehicle is moving on a curved road. The ego vehicle is approaching an intersection.} This process continues as additional positive answers refine the scene description, ensuring a coherent and progressively detailed understanding of the environment. For questions with multiple-choice answers, such as those regarding traffic signs (e.g., pedestrian crossings), predefined templates account for each possible response. By systematically structuring the output, this approach maintains clarity and completeness while efficiently conveying scene details extracted from the hierarchical question tree.

\section{Experiments and Results}

\subsection{Experiment Setup}

Experiments were conducted on a system with an RTX 4090 GPU. ROS 2 was used for real-time deployment. The fine-tuning and inference stages were carried out using the PyTorch library. \textcolor{black}{For the VLM model, we employed BLIP \cite{li2022blipbootstrappinglanguageimagepretraining} with 384.7 million trainable parameters}. The model was fine-tuned on the dataset with a learning rate of 0.000001 for 15 epochs and a batch size of 16. The use of BLIP ensures compatibility with modern common GPUs while maintaining a balance between computational efficiency, performance, and inference time for real-time applications.



\subsection{Baseline Comparison}

The fine-tuned VLM model was evaluated against the pre-trained VLM model on the proposed dataset using Lingo-Judge \cite{marcu2024lingoqavisualquestionanswering} as the evaluation metric. Lingo-Judge was selected to enable a fair comparison with the baseline model, which had not been fine-tuned on the dataset. The pre-trained VLM model, without fine-tuning, may produce answers that are semantically similar but not identical to the ground truth, leading to lower accuracy. In contrast, the fine-tuned model is expected to generate answers that closely match the ground truth, which could result in high accuracy and thus an unfair comparison. Therefore, Lingo-Judge ensures a more accurate assessment by accounting for small variations in answers, providing a consistent measure of performance.


For evaluation, a Lingo-Judge score threshold of 85\% was established, indicating a predicted answer was considered correct if it matched the ground truth with at least 85\% similarity. As shown in Table \ref{Answer_Accuracy}, the fine-tuned VLM model achieved higher accuracy across all categories, including the overall category, where it reached an accuracy of 94.81\%, compared to 72.10\% achieved by the pre-trained VLM model. This demonstrates that fine-tuning the VLM model on the driving domain and specifically adapting it to the geographical area significantly improves its performance and understanding, as compared to the baseline, which is not fine-tuned for this specific task. Results are presented in Table \ref{Answer_Accuracy}, categorized by answer type, with overall model performance summarized.

\begin{table}[h]
\caption{Accuracy Comparison Across Different Answer Categories}
\label{Answer_Accuracy}
\begin{center}
\begin{tabular}{c|c|c}
Answer Category & Fine-tuned VLM (\%)& Pre-trained VLM (\%)\\
\hline
Yes	&67.10&    52.63     \\
No	&95.65&    59.71  \\
None	&98.85&   79.53 \\
Other&	53.33&    26.66  \\
\hline

Total 	&94.81&     72.10   \\
\end{tabular}
\end{center}
\end{table}

\subsubsection{Inference Time Analysis}

Inference time was measured in milliseconds to evaluate the real-time feasibility of the hierarchical QA approach. Table \ref{Inference_Time} compares the mean inference times (with standard deviation, denoted as STD) for \textcolor{black}{the proposed hierarchical QA approach, the baseline VLM model (Without hierarchical QA), and GPT-4o.} The baseline VLM model refers to the same VLM but without employing the hierarchical question structure, indicating that it processes all 41 questions per frame, regardless of the scene content. In contrast, the hierarchical QA approach dynamically selects and answers only a subset of questions based on the scene, reducing computational overhead. This comparison highlights the effect of including the hierarchical approach and dynamically omitting unnecessary questions, improving efficiency.

For GPT, inference time was recorded from the moment the program initiates an API request until the response is received. As shown in the table \ref{Inference_Time}, GPT has a mean response time of 5 s, with a maximum of 125 s. This comparison underscores that the proposed approach is not only feasible but also cost-effective for deployment in real autonomous vehicles, where real-time processing is crucial. In contrast, using a GPT-based API for real-time scene understanding remains impractical due to its high latency.

\begin{table}[h]
\caption{Inference Time Comparison}
\label{Inference_Time}
\begin{center}
\begin{tabular}{>{\centering\arraybackslash}m{2.8cm}|>{\centering\arraybackslash}m{0.9cm}|>{\centering\arraybackslash}m{0.9cm}|>{\centering\arraybackslash}m{0.9cm}|>{\centering\arraybackslash}m{0.9cm}}
Approach & Mean (ms)& STD (ms)& Max (ms)& Min (ms)\\
\hline
Baseline VLM & 1573  & 24 &2137 &1555    \\ 
GPT-4o API &  5138  &   12253& 125970 &2719\\
Hierarchical QA (ours) & 423 & 171& 899 &309  \\
\end{tabular}
\end{center}
\end{table}

\subsection{GPT-Based Evaluation}

The scene descriptions generated by the proposed hierarchical QA approach were evaluated using the GPT-4o model, as shown in Table \ref{GPT_Comparison}. Specifically, GPT-4o assessed the generated descriptions by assigning a score from 0 to 100 based on their accuracy and completeness relative to the image. A score of 100 indicates a fully correct and comprehensive description. 
Additionally, GPT-4o was used to generate scene descriptions independently, which were then evaluated against the ground truth. To ensure a fair comparison, GPT-4o was instructed to focus solely on the specific driving-related elements present in the ground truth, including \textcolor{black}{road geometry, intersections, vehicles, pedestrians, traffic signs, speed bumps, and crosswalks}.

As shown in Table \ref{GPT_Comparison}, the proposed approach achieved a score of 65 when evaluated by GPT-4o without reference to the ground truth (GPT Score without GT). However, when compared directly with the ground truth, the score increased to 79 (GPT Score with GT). In contrast, GPT-4o, generating its own scene descriptions, achieved a score of 57 when evaluated by GPT-4o and compared to the ground truth (GPT Score with GT) and 77 when evaluated without reference (GPT Score without GT). \textcolor{black}{This comparison highlights that the proposed approach successfully demonstrates strong performance in capturing key driving scene details, while achieving a reasonable score relative to GPT-4o and remaining suitable for real-time deployment.}



\begin{table}[h]
\caption{Comparison of Driving Scene Description Accuracy}
\label{GPT_Comparison}
\begin{center}
\begin{tabular}{c|>{\centering\arraybackslash}m{1.5cm}|>{\centering\arraybackslash}m{1.5cm}}
Models & GPT Score with GT & GPT Score w/o GT \\
\hline
GPT-4o  &  57& 77 \\
Hierarchical QA (ours)   & 79 &  65\\
\end{tabular}
\end{center}
\end{table}

\subsection{Real-Time Implementation}

\begin{figure}[thpb]
      \centering

      \includegraphics[scale=0.35]{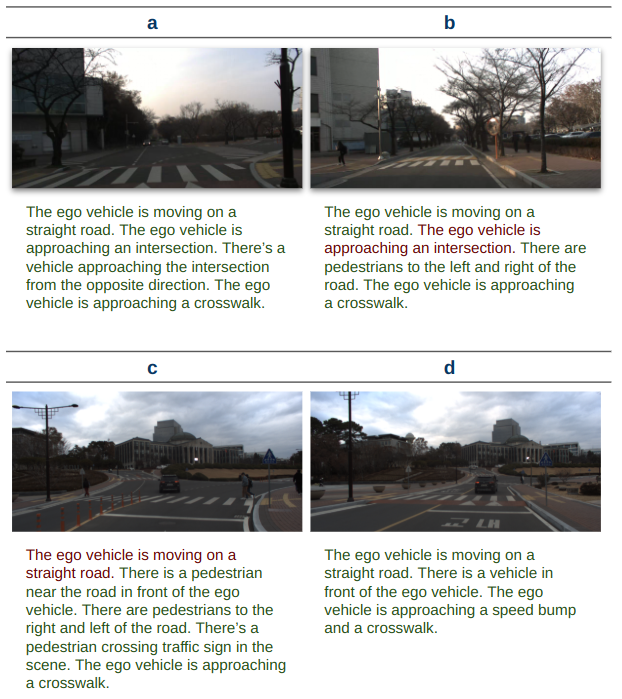}
      \caption{Driving scene understanding results from the hierarchical QA approach.}

      \label{quali1}
   \end{figure}

The hierarchical QA approach was integrated into our test vehicle using ROS 2. \textcolor{black}{Embedded as a ROS 2 node, it generated scene descriptions while the vehicle traveled around the university campus.} Despite environmental changes due to recent construction in the field test area, the proposed approach adapted effectively, demonstrating robust performance. Fig. \ref{quali1} presents some results from the hierarchical QA approach. The proposed approach successfully detects critical features for autonomous driving, such as identifying when a vehicle is approaching an intersection or when pedestrians are present at crosswalks. This information is essential to determine the course of action of the ego vehicle, particularly if the VLM were to directly issue instructions to the vehicle. The sentences in red represent incorrect descriptions, while the sentences in green highlight accurate scene understanding. In Fig. \ref{quali1}a, the proposed approach successfully detects critical information, such as a vehicle approaching the intersection from the opposite direction. In Fig. \ref{quali1}b, the proposed approach mistakenly states that the vehicle is approaching an intersection when it is already inside the intersection, but it correctly identifies pedestrians to the left and right of the crosswalk. In Fig. \ref{quali1}c, detects both pedestrians and the pedestrian crossing traffic sign. However, it fails to detect the curvature of the road. Finally, in Fig. \ref{quali1}d, it correctly identifies that there is a vehicle in front of the ego vehicle. These results demonstrate the potential of the hierarchical QA approach in enabling real-time, satisfying scene understanding for autonomous driving, although further improvements in latency and model accuracy are still necessary.


\section{Conclusion}

In this paper, we propose a hierarchical QA approach for efficient scene understanding in autonomous vehicles, offering a lightweight solution with low-latency inference. The proposed approach effectively integrates visual perception and language generation, enabling efficient scene descriptions while maintaining strong performance in interpreting dynamic driving scenarios with minimal computational cost. Furthermore, we demonstrated how the proposed approach achieves comparable scores when evaluated by GPT, indicating its effectiveness in generating accurate scene descriptions. Additionally, qualitative results from the real-time deployment show how the model dynamically captures different driving scene elements and generates coherent and contextually relevant sentences, further highlighting its potential for practical applications in autonomous driving environments.

\addtolength{\textheight}{-12cm}   


\bibliographystyle{IEEEtran}

\addtolength{\textheight}{+12cm}
\bibliography{ref.bib}

\end{document}